\documentclass[letterpaper, 10 pt, conference, onecolumn]{ieeeconf}  % Comment this line out if you need a4paper

\IEEEoverridecommandlockouts                              % This command is only needed if 
\usepackage{graphicx}
\usepackage{xcolor}
\usepackage{textcomp}

\usepackage{amsmath,amssymb,amsfonts}

\usepackage{cite}
\let\labelindent\relax 
\usepackage{enumitem}
\usepackage{multirow}
\usepackage{booktabs}
\usepackage{makecell}
\usepackage{subcaption}
\usepackage{float}
\usepackage{placeins}

\usepackage[ruled,linesnumbered]{algorithm2e}
\usepackage{soul}
\usepackage{bbding}
\title{\LARGE \bf
Heterogeneous Graph Condensation via Role-Aware Clustering
}

\author{Fuyan Ou$^{1}$ and Yulin Hu$^{2}$ and Ye Yuan$^{3}$% <-this % stops a space
}
\begin{document}

\maketitle
\thispagestyle{empty}
\pagestyle{empty}

%%%%%%%%%%%%%%%%%%%%%%%%%%%%%%%%%%%%%%%%%%%%%%%%%%%%%%%%%%%%%%%%%%%%%%%%%%%%%%%%
\begin{abstract}
Heterogeneous Graph Neural Networks (HGNNs) have exhibited remarkable efficacy in modeling complex systems with multiple types of nodes and relations, yet their training on large-scale heterogeneous graphs remains computationally prohibitive. Although graph condensation methods can effectively improve learning efficiency on large-scale graphs, existing condensation processes are mainly designed for homogeneous graphs and typically rely on computationally expensive gradient matching or bilevel optimization paradigms, rendering them impractical for heterogeneous settings. To address these limitations, we propose HGC-RC, a simple yet effective role-aware heterogeneous graph condensation framework. Specifically, HGC-RC first extracts semantically enhanced node embeddings via lightweight propagation. It then introduces a role-aware hybrid clustering strategy consisting of class-partitioned clustering for labeled target nodes to preserve class distributions and unsupervised type-wise clustering for non-target nodes to retain critical cross-type connectivity. Finally, a compact heterogeneous graph is efficiently reconstructed based on the resulting cluster assignments. Extensive experiments demonstrate that HGC-RC outperforms state-of-the-art baselines, offering a practical pathway to accelerate HGNN training on large-scale heterogeneous graphs without sacrificing task performance.
\end{abstract}

\begin{keywords}
Heterogeneous graph condensation, heterogeneous graph neural networks, role-aware clustering, graph reduction
\end{keywords}

%%%%%%%%%%%%%%%%%%%%%%%%%%%%%%%%%%%%%%%%%%%%%%%%%%%%%%%%%%%%%%%%%%%%%%%%%%%%%%%%

\section{INTRODUCTION}
Heterogeneous graphs are widely used to model complex relational systems in real-world scenarios, such as academic networks, recommender systems, and knowledge-driven platforms~\cite{yuan2020generalized, tang2025autoencoding, liao2025proximal, hu2025comprehensive, gou2026multiscale, qin2024adaptively}. Prior to the rise of heterogeneous models, standard Graph Neural Networks (GNNs)~\cite{GCN, GAT, GraphSAGE, Fast-GCN, GraphSAINT, yuan2025node, he2025modularized, bi2025graph, li2024generalized} demonstrated immense potential in homogeneous graph representation learning. By explicitly characterizing multiple node and relation types, heterogeneous graphs provide richer semantics than homogeneous graphs. To effectively model such heterogeneous interactions, heterogeneous graph neural networks (HGNNs)~\cite{wang2026graph,wang2026advanced,he2026tensor,HAN,HetGNN,MAGNN,HetSANN,Nars, wang2025gt, lin2026ncsac, yang2025link, wu2025graph, bi2024fast} have achieved strong performance in a variety of domains, including traffic networks~\cite{yuan2026novel,qin2026robust, yuan2020temporal, xu2026sampling, liao2025novel, xu2025adaptively, hou2025multiaspect, chen2025latent, qin2024asynchronous}, biology~\cite{bi2025discovering,han2025sgd, li2025knowledge, yang2025fmvpci, deng2026fuzzy, luo2025analysis, wu2025multi, liu2024symmetry}, relational databases~\cite{wu2025learning,lyu2026dynamic,li2026adaptive, wu2026nongradient, yuan2023kalman, li2026neural, wu2025outlier, wu2024prediction}, as well as emerging applications like event detection, link prediction, and disease prognosis~\cite{dse1,dse2,jcst1,jcst2,scis1,scis2, wang2025convolution, he2025structure, chen2024generalized}. Despite their expressive power, training HGNNs on large heterogeneous graphs remains expensive due to repeated relation-aware propagation, high-dimensional semantic features, and neighborhood expansion.

\emph{Graph condensation} (GC), which constructs a compact graph while preserving downstream utility, has emerged as a promising paradigm for accelerating GNN training~\cite{gcond, li2025learning, yuan2024fuzzy, yuan2024adaptive, lyu2025genetic, luo2025calibrator, qin2024parallel}. Existing GC methods have achieved encouraging results on homogeneous graphs, where most approaches rely on gradient matching or bi-level optimization to learn compact synthetic structures and features~\cite{yuan2023adaptive, yuan2022multilayered, xu2025attention, wei2024robust}. However, extending these methods to heterogeneous settings is far from straightforward. In heterogeneous graphs, multiple feature spaces, relation semantics, and cross-type dependencies must be preserved simultaneously, while naive homogenization, such as relation union or meta-path projection, may distort relation semantics and class balance. Although a few studies have begun to consider heterogeneous graph condensation, such as HGCond~\cite{HGCond}, these methods remain optimization-heavy and sensitive to the relay model.

\begin{figure}[!t]
    \centering
    \includegraphics[width=\linewidth]{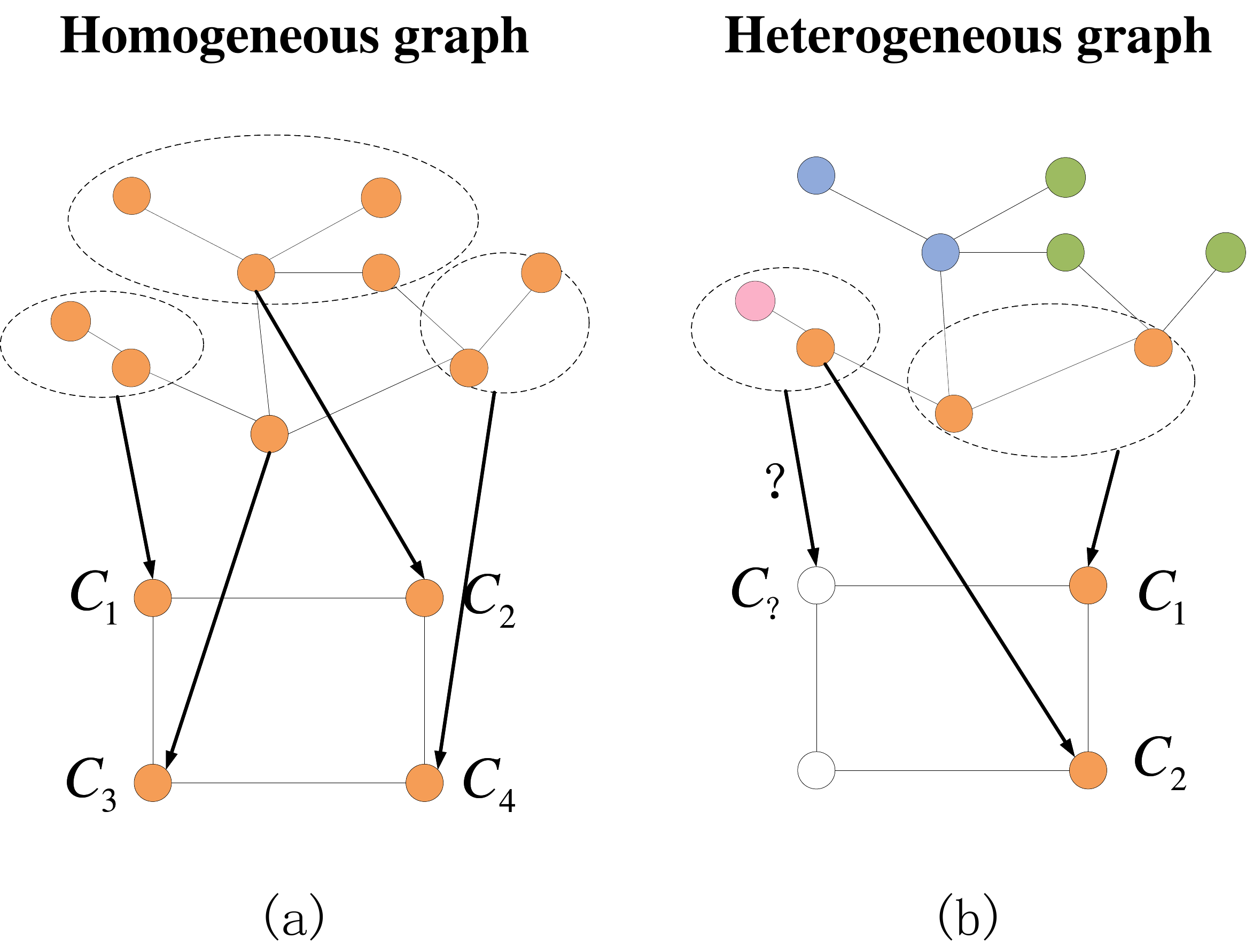}
    \caption{Motivation of role-aware condensation in heterogeneous graphs. (a) Uniform condensation suits homogeneous graphs. (b) In heterogeneous graphs, it may mix target classes and remove support nodes.}
    \label{fig:1}
    \vspace{-0.5cm}
\end{figure}

A key challenge of heterogeneous graph condensation lies in the \emph{role asymmetry} among node types. Specifically, labeled \emph{target-type} nodes directly determine the downstream decision boundary, whereas \emph{non-target} nodes mainly provide semantic and structural support. As illustrated in Figure~\ref{fig:1}, applying a uniform condensation strategy may be suitable for homogeneous graphs, where nodes play relatively consistent roles. In contrast, in heterogeneous graphs, treating all nodes equally during condensation may blur the class structure of target nodes and discard critical support nodes. Therefore, capturing and exploiting node-role differences is essential for effective heterogeneous graph condensation.

To address this issue, we propose \textbf{HGC-RC}, a role-aware framework for heterogeneous graph condensation. Instead of relying on expensive iterative synthetic graph optimization, HGC-RC constructs a compact heterogeneous graph from a data-centric perspective. Specifically, HGC-RC first computes semantic embeddings using SeHGNN preprocessing, then performs class-partitioned clustering for target nodes and type-wise unsupervised clustering for non-target nodes. Based on the resulting cluster assignments, it aggregates original node features and reconstructs a compact heterogeneous graph through inter-cluster connectivity summarization.

Our main contributions are as follows:
\begin{itemize}
\item We formulate heterogeneous graph condensation from a \emph{role-aware} perspective, highlighting the asymmetric roles of target and non-target nodes in preserving downstream classification utility.
\item We propose \textbf{HGC-RC}, an efficient optimization-free framework that combines class-partitioned clustering for target nodes, type-wise unsupervised clustering for non-target nodes, and cluster-level graph reconstruction.
\item Experiments on three heterogeneous benchmarks show that \textbf{HGC-RC} achieves competitive performance under high compression with substantially lower condensation cost than representative optimization-based baselines.
\end{itemize}

\section{PRELIMINARIES}
In this section, we introduce the notations, problem formulation, and background on graph condensation.

\subsection{Notations and Problem Formulation}\label{sec:problem}
A heterogeneous graph is denoted as 
$\mathcal{G}=(\mathcal{V},\mathcal{E},\phi,\psi)$, 
where $\mathcal{V}$ and $\mathcal{E}$ are the node and edge sets, respectively.
The mapping $\phi:\mathcal{V}\rightarrow\mathcal{T}$ assigns each node to a node type, and
$\psi:\mathcal{E}\rightarrow\mathcal{R}$ assigns each edge to a relation type,
where $\mathcal{T}$ and $\mathcal{R}$ denote the sets of node types and relation types, respectively.
The graph is homogeneous when $|\mathcal{T}|=|\mathcal{R}|=1$.

We focus on a node classification task defined on a \emph{target node type} $\tau_t\in\mathcal{T}$ (e.g., \textit{Author} in DBLP~\cite{HGB}), while the remaining node types act as \emph{non-target} types that provide semantic context and structural support.
Let
\begin{equation}
\mathcal{V}_t=\{v\in\mathcal{V}\mid \phi(v)=\tau_t\}
\end{equation}
be the target-type node set, and let
\begin{equation}
\mathcal{V}_u=\mathcal{V}\setminus\mathcal{V}_t
\end{equation}
be the non-target node set.
Target nodes have features $\mathbf{X}_{\tau_t}$ and labels $\mathbf{y}\in\{1,\dots,C\}^{|\mathcal{V}_t|}$ over $C$ classes, whereas non-target nodes have features $\{\mathbf{X}_{\tau}\}_{\tau\in\mathcal{T}\setminus\{\tau_t\}}$ and no task labels.

\noindent\textbf{Meta-path features.}
Following common heterogeneous learning practice, we consider a set of meta-paths $\mathcal{P}$ defined over node types.
A meta-path $P\in\mathcal{P}$ can be written as
\begin{equation}
P:\tau_1\rightarrow\tau_2\rightarrow\cdots\rightarrow\tau_L.
\end{equation}
For each target node, meta-path-based features can be constructed by propagating information along $P$, yielding $\mathbf{X}^{P}$.

\noindent\textbf{Problem definition.}
Given the original heterogeneous graph $\mathcal{G}$ and a condensation ratio $\rho\in(0,1)$, our goal is to construct a condensed heterogeneous graph
\begin{equation}
\mathcal{G}_c=\big(\mathcal{V}_c,\{\mathbf{A}_r^c\}_{r\in\mathcal{R}},\mathbf{X}_c,\mathbf{y}_c\big),
\end{equation}
with $|\mathcal{V}_c|\ll|\mathcal{V}|$, such that an HGNN trained on $\mathcal{G}_c$ achieves performance comparable to that obtained on $\mathcal{G}$.
We allocate a type-wise condensation budget:
\begin{equation}
|\mathcal{V}_c^{(\tau)}|
=
\max\left(1,\left\lfloor \rho\,|\mathcal{V}^{(\tau)}|\right\rfloor\right),
\end{equation}
where
\begin{equation}
\mathcal{V}^{(\tau)}=\{v\in\mathcal{V}\mid \phi(v)=\tau\}
\end{equation}
is the node set of type $\tau$, and $\mathcal{V}_c^{(\tau)}$ denotes the condensed nodes of type $\tau$.
For the target type, HGC-RC further allocates the budget across classes within $\mathcal{M}$ to preserve class balance.
The condensed features $\mathbf{X}_c$ are obtained by aggregating original node features according to the cluster assignments learned in the embedding space.

Formally, let $f_\theta$ be an HGNN and let $\mathcal{L}$ denote the task loss on target nodes.
Our goal is to construct a condensed graph such that training an HGNN on $\mathcal{G}_c$ preserves the downstream utility of training on the original graph $\mathcal{G}$ under the same task, while satisfying the budget controlled by $\rho$.
In practice, $\mathcal{G}_c$ serves as a compact \emph{training graph}.
The condensation mask $\mathcal{M}$ corresponds to the labeled training target nodes to be condensed, while validation and test target nodes remain on the original graph for downstream evaluation.
Therefore, the objective of condensation is to preserve the training utility of the original heterogeneous graph under a strict node budget without changing the standard evaluation protocol.

\subsection{Graph Reduction}
Graph reduction aims to reduce graph size while preserving downstream GNN utility. Existing methods mainly fall into three categories: \emph{graph sparsification}, \emph{graph coarsening}, and \emph{graph condensation}~\cite{survey-graph-reduction, yuan2025proportional, liu2025scalable, jiang2024iterative}.

\textbf{Graph Sparsification \& Coarsening.}
Graph sparsification selects a subset of nodes or edges to approximately preserve the original graph quality~\cite{Sparsification1, bi2025spatiotemporal}. Representative coreset-based methods include Herding~\cite{coreset1} and K-center~\cite{coreset2}, which select representative samples in the feature space. In heterogeneous graphs, sparsification is more challenging because node/edge types and meta-path semantics must be preserved; naive sampling may distort relation distributions and class balance, causing notable accuracy drops at low keep ratios~\cite{HSparsification}. Graph coarsening groups original nodes into super-nodes and defines their connections to preserve structural information~\cite{graph-coarsening}. However, under high compression, sparsification and coarsening may lose critical semantics and are often too topology-centered to reliably support heterogeneous downstream tasks.

\textbf{Graph Condensation.}
Graph condensation (GC) compresses a large graph into a much smaller \emph{synthetic} graph while preserving training utility, inspired by the success of dataset distillation in computer vision~\cite{Dataset-Condensation1,Dataset-Condensation2,Dataset-Condensation4}. A representative early method, GCond~\cite{gcond}, formulates GC as a bi-level optimization problem that matches gradients between the original and synthetic graphs to jointly learn condensed features and structure. Subsequent methods improve efficiency by avoiding costly inner-loop optimization and modeling discrete graph structures probabilistically~\cite{one-step}. Other directions include trajectory-based methods, structure-free condensation such as SFGC~\cite{SFGC}, spectrum-oriented regularization such as eigenbasis matching~\cite{Eigenbasis-Matching, GCEM}, and distribution matching techniques~\cite{Receptive-Field-Distribution-Matching}. More recent GC methods further improve efficiency through precomputed features, simplified optimization, and lower construction cost~\cite{gao1,gao2,gao3,Bonsai,GCGP, wang2024distributed}, among which GCPA~\cite{gcpa} is representative. Nevertheless, most GC methods remain optimization-intensive and are mainly designed for homogeneous graphs.

\begin{figure*}[!htbp]
\centerline{\includegraphics[width=7.0in]{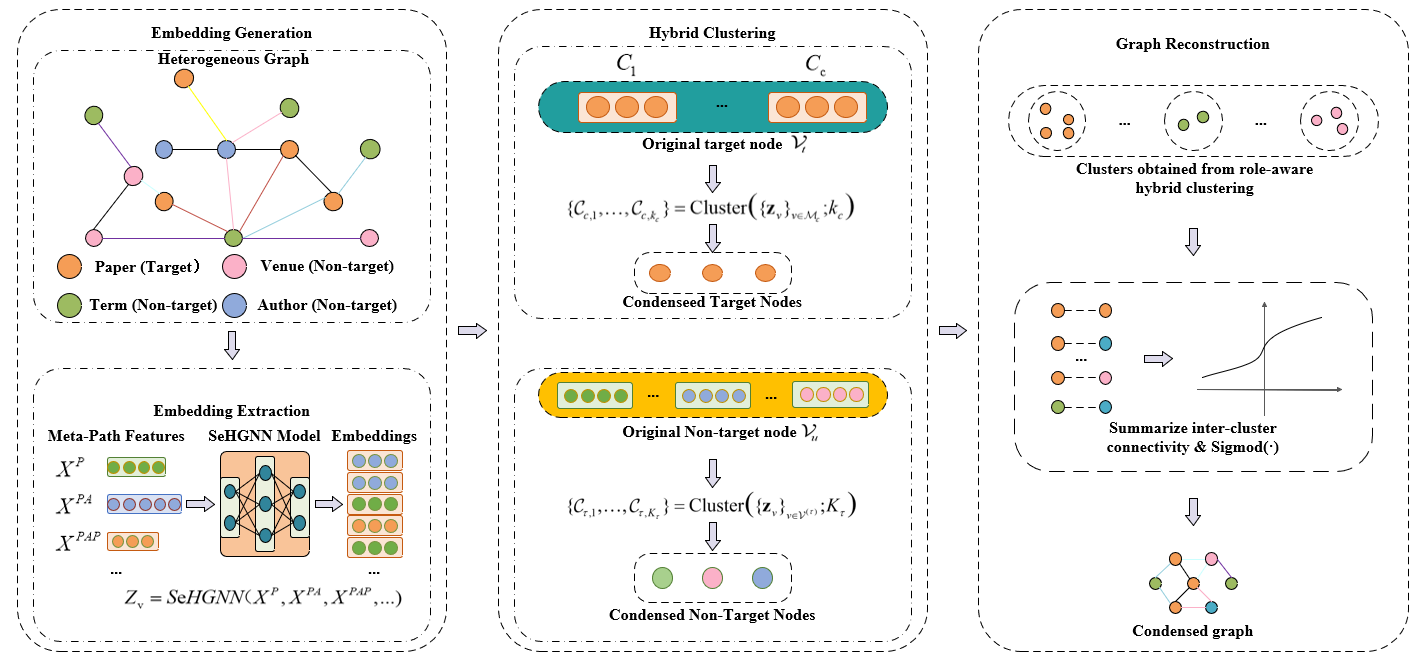}} 
\caption{The workflow of HGC-RC.}
\label{fig:2}
\vspace{-0.5cm}
\end{figure*}

\textbf{Heterogeneous Graph Condensation.}
Extending GC to heterogeneous graphs is non-trivial due to multi-type feature spaces, relation-specific dependencies, and the need to preserve heterogeneous semantics. HGCond~\cite{HGCond} is among the first methods for this setting; it leverages clustering for initialization and adopts an orthogonal parameter sequence strategy to improve optimization stability. However, it still relies on optimization-driven condensation and is sensitive to the relay model and budget settings. Graph-Skeleton~\cite{graph-skeleton} also considers heterogeneous graphs, but condenses only background nodes while keeping all target nodes, which differs from general settings that compress both target and non-target parts.

\section{FRAMEWORK}
In this section, we present \textbf{HGC-RC}, a role-aware framework for heterogeneous graph condensation. As shown in Figure~\ref{fig:2}, HGC-RC consists of semantic embedding extraction, role-aware clustering, and cluster-level graph reconstruction, aiming to preserve discriminative target information and structural-semantic support in a compact training graph.

\subsection{Semantic-Enhanced Embedding Generation}
Given $\mathcal{G}=(\mathcal{V},\mathcal{E},\phi,\psi)$ and a set of meta-paths $\mathcal{P}$, HGC-RC first computes semantic-enhanced embeddings for all nodes to provide a unified representation space for condensation.
Specifically, for each meta-path $P\in\mathcal{P}$, we construct meta-path features for target nodes by propagating information along $P$, yielding $\mathbf{X}^{P}$.

Let $\mathbf{Z}\in\mathbb{R}^{|\mathcal{V}|\times d}$ denote the embedding matrix, where the row vector $\mathbf{z}_v$ corresponds to node $v$.
We write the embedding extraction stage as
\begin{equation}
\mathbf{Z} = g\Big(\mathcal{G},\{\mathbf{X}^{P}\}_{P\in\mathcal{P}}\Big),
\end{equation}
where $g(\cdot)$ is instantiated by a SeHGNN~\cite{SeHGNN} preprocessing encoder to capture multi-relational semantics with low overhead.
The resulting embeddings encode both structural context and semantic patterns induced by meta-paths, and serve as the unified input to the subsequent role-aware hybrid clustering stage.

\subsection{Role-aware Hybrid Clustering}
HGC-RC performs role-aware hybrid clustering in the embedding space $\mathbf{Z}$.
The key idea is that \emph{target nodes} should preserve class-discriminative structure, while \emph{non-target nodes} should preserve structural support and cross-type connectivity.
Accordingly, HGC-RC applies class-partitioned clustering to target nodes and unsupervised type-wise clustering to non-target nodes, thereby obtaining a mapping from original nodes to condensed clusters.

\noindent\textbf{Target nodes: class-partitioned clustering.}
Let $\mathcal{M}\subseteq\mathcal{V}_t$ denote the condensation mask (typically the labeled training target nodes), and labels $\mathbf{y}$ are only used within $\mathcal{M}$.
Given ratio $\rho$, we allocate the target-type budget
\begin{equation}
K_t=\max\Big(1,\big\lfloor \rho\,|\mathcal{M}|\big\rfloor\Big).
\end{equation}
To preserve class balance, we further distribute $K_t$ across classes according to the label distribution in $\mathcal{M}$.
For class $c\in\{1,\dots,C\}$, let
\begin{equation}
\mathcal{M}_c=\{v\in\mathcal{M}\mid y_v=c\},
\end{equation}
where $y_v$ denotes the label of node $v$.
We allocate
\begin{equation}
k_c=\max\Big(1,\Big\lfloor K_t\cdot\frac{|\mathcal{M}_c|}{|\mathcal{M}|}\Big\rfloor\Big),
\quad \text{and adjust } \sum_{c=1}^{C}k_c=K_t.
\end{equation}
We then cluster embeddings within each class:
\begin{equation}
\{\mathcal{C}_{c,1},\dots,\mathcal{C}_{c,k_c}\}
=\mathrm{Cluster}\Big(\{\mathbf{z}_v\}_{v\in\mathcal{M}_c};\,k_c\Big),
\end{equation}
where $\mathrm{Cluster}(\cdot)$ is instantiated by spectral clustering or k-means.
This defines a mapping from target nodes in $\mathcal{M}$ to condensed target clusters. Nodes outside $\mathcal{M}$ do not participate in target-node condensation. Since target-node clustering is performed in a class-partitioned manner, each condensed target cluster is assigned the corresponding class label $c$.

\noindent\textbf{Non-target nodes: unsupervised clustering.}
For each non-target type $\tau\in\mathcal{T}\setminus\{\tau_t\}$, let
\begin{equation}
\mathcal{V}^{(\tau)}=\{v\in\mathcal{V}\mid \phi(v)=\tau\}.
\end{equation}
We allocate a type-wise budget consistent with the problem formulation:
\begin{equation}
K_{\tau}=\max\Big(1,\big\lfloor \rho\,|\mathcal{V}^{(\tau)}|\big\rfloor\Big).
\end{equation}
We then cluster embeddings of this type without using labels:
\begin{equation}
\{\mathcal{C}_{\tau,1},\dots,\mathcal{C}_{\tau,K_{\tau}}\}
=\mathrm{Cluster}\Big(\{\mathbf{z}_v\}_{v\in\mathcal{V}^{(\tau)}};\,K_{\tau}\Big),
\end{equation}
thereby defining a mapping from original non-target nodes to condensed non-target clusters. These clusters preserve structural support and cross-type semantic context for downstream learning.

Finally, the condensed node set is written as
\begin{equation}
\mathcal{V}_c=\mathcal{V}_{c}^{(t)}\cup \mathcal{V}_{c}^{(u)},
\end{equation}
where $\mathcal{V}_{c}^{(t)}$ consists of all condensed clusters obtained from target nodes in $\mathcal{M}$, and $\mathcal{V}_{c}^{(u)}$ consists of all condensed clusters obtained from non-target node types.
By construction, $\mathbf{y}_c$ is defined only on $\mathcal{V}_{c}^{(t)}$, while non-target condensed nodes have no task labels.

\subsection{Condensed Graph Reconstruction}
Let $s(v)$ denote the cluster assignment of node $v$. Based on the resulting assignments, HGC-RC reconstructs a condensed heterogeneous graph by aggregating node features and summarizing inter-cluster connectivity.

\noindent\textbf{Condensed feature construction.}
Let $\mathbf{x}_v$ denote the original feature of node $v$, and let
\begin{equation}
\mathcal{C}_k=\{v\mid s(v)=k\}
\end{equation}
be the set of nodes assigned to condensed cluster $k$. The condensed feature is computed by
\begin{equation}
\mathbf{x}^{c}_k=\frac{1}{|\mathcal{C}_k|}\sum_{v\in\mathcal{C}_k}\mathbf{x}_v.
\end{equation}
For target nodes, only nodes in the condensation mask participate in feature aggregation.

\noindent\textbf{Condensed label construction.}
Labels are defined only for condensed target clusters, each of which inherits the corresponding class label.

\noindent\textbf{Condensed adjacency construction.}
For each relation type $r\in\mathcal{R}$, let
\begin{equation}
\mathcal{E}_r=\{e=(u,v)\in\mathcal{E}\mid \psi(e)=r\}
\end{equation}
denote the edge set of relation $r$. We first compute the inter-cluster edge count matrix:
\begin{equation}
W_r^c(i,j)=\sum_{(u,v)\in\mathcal{E}_r}\mathbf{1}[s(u)=i,\ s(v)=j].
\end{equation}
To reduce cluster-size bias, we normalize it as
\begin{equation}
\widetilde{W}_r^c(i,j)=\frac{W_r^c(i,j)}{|\mathcal{C}_i|\,|\mathcal{C}_j|+\epsilon},
\end{equation}
where $\epsilon$ is a small constant. We then compute a bounded edge score:
\begin{equation}
S_r^c(i,j)=\sigma\!\left(\alpha\big(\widetilde{W}_r^c(i,j)-\beta_r\big)\right),
\end{equation}
where $\sigma(\cdot)$ is the sigmoid function, $\alpha$ controls the sharpness, and $\beta_r$ is a relation-specific threshold. The condensed adjacency is defined as
\begin{equation}
A_r^c(i,j)=\mathbf{1}\!\left[S_r^c(i,j)>\delta\right],
\end{equation}
where $\delta$ is a filtering threshold. In our implementation, $\alpha=10$, $\delta=0.5$, and $\beta_r$ is set as the mean of nonzero $\widetilde{W}_r^c(i,j)$ values for relation $r$. The condensed graph is written as
\begin{equation}
\mathcal{G}_c=\Big(\mathcal{V}_c,\{\mathbf{A}_r^{c}\}_{r\in\mathcal{R}},\mathbf{X}_c,\mathbf{y}_c\Big).
\end{equation}

% \begin{algorithm}[t]
% {
% \small
% \caption{HGC-RC: Role-aware Heterogeneous Graph Condensation}
% \label{alg:hgcrc}
% \SetKwInOut{Input}{Input}\SetKwInOut{Output}{Output}
% \Input{Heterogeneous graph $\mathcal{G}$; target node type $\tau_t$; mask $\mathcal{M}$; labels $\mathbf{y}$; ratio $\rho$; meta-path set $\mathcal{P}$.}
% \Output{Condensed graph $\mathcal{G}_c$.}

% $\mathbf{Z}\leftarrow g\big(\mathcal{G},\{\mathbf{X}^{P}\}_{P\in\mathcal{P}}\big)$\\
% $K_t\leftarrow \max(1,\lfloor \rho|\mathcal{M}|\rfloor)$; compute class budgets $\{k_c\}_{c=1}^{C}$\\
% \For{$c=1$ \KwTo $C$}{
%     cluster $\{\mathbf{z}_v\}_{v\in\mathcal{M}_c}$ into $k_c$ clusters and assign nodes to target clusters\\
% }
% \For{$\tau\in\mathcal{T}\setminus\{\tau_t\}$}{
%     $K_{\tau}\leftarrow \max(1,\lfloor \rho|\mathcal{V}^{(\tau)}|\rfloor)$\\
%     cluster $\{\mathbf{z}_v\}_{v\in\mathcal{V}^{(\tau)}}$ into $K_{\tau}$ clusters and assign nodes to non-target clusters\\
% }
% derive unified assignment $s(v)$ and condensed labels $\mathbf{y}_c$\\
% aggregate original node features to obtain $\mathbf{X}_c$\\
% \For{$r\in\mathcal{R}$}{
%     compute $\mathbf{W}_r^c$, normalize it to $\widetilde{\mathbf{W}}_r^c$, and obtain $\mathbf{A}_r^c$ by sigmoid scoring and threshold filtering\\
% }
% construct $\mathcal{G}_c=(\mathcal{V}_c,\{\mathbf{A}_r^c\}_{r\in\mathcal{R}},\mathbf{X}_c,\mathbf{y}_c)$\\
% \textbf{Return:} {$\mathcal{G}_c$}
% }
% \end{algorithm}

\begin{table*}[!htbp]
\renewcommand\arraystretch{1.2}
\caption{\textbf{Experiment results of node classification prediction tasks on three datasets.}}
\centering
\resizebox{0.9\textwidth}{!}{
\begin{tabular}{*{10}{c}}%
\toprule%
& &\multicolumn{6}{c}{Baselines}& Proposed\\
\cmidrule(r){3-8} \cmidrule(r){9-9}
Dataset&Ratio (r)&Random-HG&K-Center-HG&Coarsening-HG&GCond&GCPA&HGCond&\pmb{HGC-RC}&Whole Dataset\\
\hline
\multirow{4}{*}{ACM}
&
1.2\%&$54.21\pm1.92$&$63.54\pm1.73$&$63.29\pm2.05$&$41.78\pm8.62$&$68.62\pm0.41$&$87.41\pm1.98$&$\pmb{90.76\pm0.38}$&\multirow{4}{*}{$94.18\pm0.14$}\\
&
2.4\%&$60.12\pm2.11$&$66.23\pm1.67$&$64.69\pm1.88$&$48.59\pm6.28$&$69.71\pm0.52$&$88.21\pm2.02$&$\pmb{91.26\pm0.68}$&\\
&
4.8\%&$60.87\pm1.79$&$68.72\pm2.14$&$69.53\pm1.66$&$50.25\pm4.56$&$69.73\pm0.46$&$84.35\pm1.90$&$\pmb{91.74\pm0.40}$&\\
&
9.6\%&$65.41\pm1.74$&$76.54\pm2.18$&$71.88\pm1.82$&$45.36\pm6.32$&$68.61\pm0.39$&$84.21\pm1.69$&$\pmb{93.46\pm0.23}$&\\
\hline
\multirow{4}{*}{DBLP}
&
1.2\%&$39.65\pm2.07$&$60.41\pm1.95$&$54.33\pm1.71$&$53.49\pm5.54$&$76.41\pm0.88$&$\pmb{92.34\pm1.88}$&$\underline{88.94\pm0.68}$&\multirow{4}{*}{$95.18\pm0.18$}\\
&
2.4\%&$49.58\pm1.83$&$64.92\pm2.06$&$58.71\pm1.64$&$44.95\pm4.27$&$75.17\pm0.96$&$\pmb{93.86\pm2.12}$&$\underline{89.92\pm1.07}$&\\
&
4.8\%&$44.63\pm1.76$&$71.54\pm2.19$&$65.33\pm1.97$&$52.95\pm9.23$&$76.24\pm0.83$&$\underline{91.62\pm1.81}$&$\pmb{91.70\pm0.69}$&\\
&
9.6\%&$56.93\pm2.15$&$78.74\pm1.69$&$77.66\pm2.02$&$46.23\pm8.25$&$75.79\pm0.79$&$\underline{91.47\pm1.75}$&$\pmb{92.79\pm0.36}$&\\
\hline
\multirow{4}{*}{IMDB}
&
1.2\%&$40.16\pm1.87$&$42.71\pm2.10$&$41.26\pm1.78$&$42.12\pm6.57$&$34.52\pm0.31$&$56.54\pm1.93$&$\pmb{60.14\pm0.42}$&\multirow{4}{*}{$69.84\pm0.24$}\\
&
2.4\%&$41.69\pm2.17$&$50.55\pm1.61$&$44.72\pm1.99$&$38.54\pm3.28$&$34.56\pm0.29$&$59.24\pm2.05$&$\pmb{62.81\pm0.36}$&\\
&
4.8\%&$49.12\pm1.72$&$52.86\pm2.20$&$49.97\pm1.65$&$32.75\pm5.36$&$34.50\pm0.27$&$56.79\pm1.84$&$\pmb{63.86\pm0.46}$&\\
&
9.6\%&$50.19\pm2.03$&$59.42\pm1.86$&$49.87\pm2.16$&$40.56\pm7.96$&$34.59\pm0.33$&$58.58\pm1.70$&$\pmb{65.74\pm0.17}$&\\
\hline
\end{tabular}}
\label{exp1}
\end{table*}

\section{EXPERIMENTS}
\subsection{Experimental Setup}

\begin{table}[t]
\renewcommand\arraystretch{1}
\caption{\textbf{Overview of the datasets.}}
\setlength{\tabcolsep}{1mm}
\centering
\resizebox{0.45\textwidth}{!}{
\begin{tabular}{*{7}{c}}%
\hline% 
\makecell[c]{Datasets}&\#Nodes&\makecell[c]{\#Nodes \\types}&\#Edges&\makecell[c]{\#Edge \\types}&Target&\#Classes\\
\hline% 
DBLP&26,128&4&239,566&6&author&4\\
ACM&10,942&4&547,872&8&paper&3\\
IMDB&21,420&4&86,642&6&movie&5\\
\hline
\end{tabular}}
\label{tab:Datasets}
\vspace{-0.5cm}
\end{table}

\subsubsection{Datasets}
We evaluate HGC-RC on three widely used heterogeneous graph benchmarks: \textbf{ACM}~\cite{HAN}, \textbf{DBLP}~\cite{HGB}, and \textbf{IMDB}~\cite{HGB}. Following the standard HGB setting, \emph{paper} nodes in ACM, \emph{author} nodes in DBLP, and \emph{movie} nodes in IMDB are treated as target node types, with node labels split into 24\%, 6\%, and 70\% for training, validation, and testing, respectively. We evaluate all methods under four condensation ratios, i.e., $1.2\%$, $2.4\%$, $4.8\%$, and $9.6\%$.

\subsubsection{Baselines}
We compare HGC-RC with representative graph reduction and condensation baselines, including \textbf{Random-HG}, \textbf{K-Center-HG}, \textbf{Coarsening-HG}, \textbf{GCond}, \textbf{GCPA}, and \textbf{HGCond}. Random-HG randomly samples nodes under the same budget. K-Center-HG selects representative nodes in the embedding space. Coarsening-HG applies topology-based graph coarsening. GCond and GCPA are homogeneous graph condensation methods transferred to the heterogeneous setting through a homogenized graph view for fair comparison. HGCond is a heterogeneous-specific condensation baseline. All methods are evaluated under the same condensation ratios and downstream task.

\subsubsection{Implementation Details}
We use SeHGNN preprocessing to generate semantic-enhanced embeddings and then perform role-aware clustering in the embedding space. For SeHGNN, the hidden dimension and embedding size are both set to 512. We use 2 propagation layers for feature preprocessing, while the number of task layers is set to 3, 1, and 4 on DBLP, ACM, and IMDB, respectively. The dropout rate is 0.5 for all datasets, and residual connections are enabled on DBLP. We train the embedding model with Adam, using learning rate 0.001 and weight decay 0, for 200 epochs with early-stopping patience 50. Unless otherwise specified, the same condensation ratio is applied across node types, and spectral clustering is used by default. For downstream evaluation, we adopt SeHGNN as the HGNN backbone and report mean accuracy and standard deviation over 5 independent runs. For condensed adjacency construction, we set $\alpha=10$, $\delta=0.5$, and define the relation-specific threshold $\beta_r$ as the mean of nonzero normalized inter-cluster connectivity scores for relation $r$.

\subsection{Main Results}

Table~\ref{exp1} reports node classification accuracy on ACM, DBLP, and IMDB under different condensation ratios. Overall, HGC-RC achieves competitive performance across all settings and the best results in most of them, indicating that the proposed role-aware condensation strategy preserves substantial downstream utility under severe compression.

On \textbf{ACM}, HGC-RC performs best across all evaluated ratios and remains relatively close to the full-graph performance, suggesting that class-partitioned target condensation preserves discriminative information effectively. On \textbf{DBLP}, HGC-RC is weaker than HGCond at the two lowest ratios, but becomes slightly better at higher ratios, which suggests that the proposed clustering-based condensation benefits from a less restrictive budget. On \textbf{IMDB}, HGC-RC achieves the best performance across all ratios, highlighting the usefulness of preserving supportive non-target semantics in heterogeneous graphs with complex interactions.

Overall, HGC-RC provides a favorable accuracy-compression trade-off while avoiding expensive iterative synthetic graph optimization.

\begin{figure}[t!]
\centering
\includegraphics[width=3.3in]{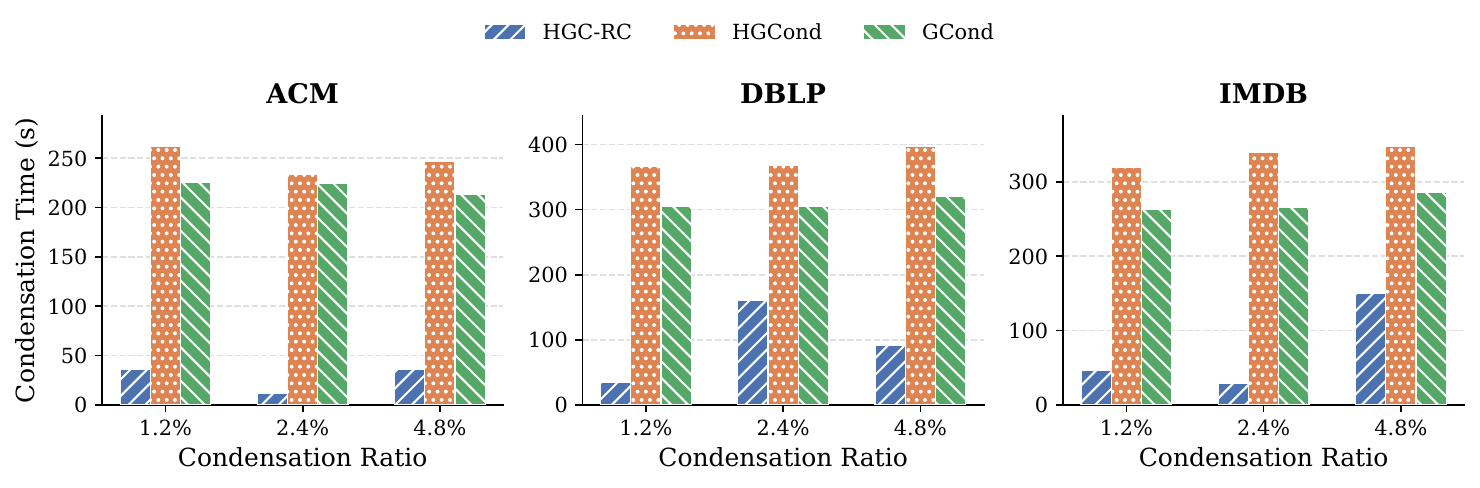}
\caption{Runtime comparison with other methods.}
\label{fig:time_cost_analysis}
\vspace{-0.8cm}
\end{figure}

\subsection{Efficiency Analysis}

Figure~\ref{fig:time_cost_analysis} compares the condensation time of HGC-RC with HGCond and GCond on ACM, DBLP, and IMDB. HGC-RC is consistently more efficient than both optimization-based baselines across datasets and condensation ratios, especially under larger budgets.

This advantage comes from avoiding repeated gradient matching and synthetic graph optimization. Instead, HGC-RC relies on a one-shot pipeline of semantic embedding extraction, role-aware clustering, and direct graph reconstruction. Although the absolute runtime varies across datasets, HGC-RC maintains a clear condensation-time advantage in all evaluated cases.

\begin{table}[!htbp]
\centering
\caption{\textbf{Ablation study at 2.4\% condensation ratio.} $\Delta$ denotes the performance change (Variant$-$HGC-RC) on the same dataset.}
\label{tab:ablation_24}
\renewcommand\arraystretch{1.1}
\setlength{\tabcolsep}{6pt}
\begin{tabular}{l l c c}
\toprule
\textbf{Dataset} & \textbf{Variant} & \textbf{Accuracy (\%)} & \textbf{$\Delta$} \\
\midrule
\multirow{3}{*}{IMDB} 
& \textbf{HGC-RC} & $62.81 \pm 0.36$ & -- \\
& w/o semantic embedding & $59.28 \pm 0.45$ & $-3.53$ \\
& w/o class partition & $59.66 \pm 0.55$ & $-3.15$ \\
\midrule
\multirow{3}{*}{DBLP} 
& \textbf{HGC-RC} & $89.92 \pm 1.07$ & -- \\
& w/o semantic embedding & $82.11 \pm 0.91$ & $-7.81$ \\
& w/o class partition & $86.03 \pm 0.49$ & $-3.89$ \\
\midrule
\multirow{3}{*}{ACM} 
& \textbf{HGC-RC} & $91.26 \pm 0.68$ & -- \\
& w/o semantic embedding & $89.89 \pm 0.17$ & $-1.37$ \\
& w/o class partition & $88.20 \pm 2.82$ & $-3.06$ \\
\bottomrule
\end{tabular}
\vspace{-0.3cm}
\end{table}

\subsection{Ablation Study}
We conduct ablation studies at the $2.4\%$ condensation ratio on ACM, DBLP, and IMDB to evaluate two key components: semantic-enhanced embedding and class-partitioned target clustering. Table~\ref{tab:ablation_24} reports the results, where $\Delta$ denotes the performance gap between each variant and the full HGC-RC model.

Removing semantic-enhanced embedding consistently degrades performance on all datasets, with the largest drop observed on DBLP. This suggests that semantic preprocessing organizes heterogeneous nodes into a more informative representation space for subsequent clustering.

Removing class partition also causes clear drops across all datasets, indicating that target-node condensation should preserve label-aware structure rather than treating all target nodes uniformly. In particular, class-partitioned clustering helps maintain class balance and reduces the risk of merging semantically different classes into overly coarse condensed target nodes.

Overall, the ablation results support the two main design choices in HGC-RC: semantic-enhanced embedding provides a more informative clustering space, and role-aware target condensation improves discriminative preservation under high compression.

\section{CONCLUSIONS}
This paper presents \textbf{HGC-RC}, an efficient role-aware framework for heterogeneous graph condensation. By combining semantic-enhanced embedding extraction, hybrid clustering for target and non-target nodes, and cluster-level graph reconstruction, HGC-RC produces compact training graphs without costly iterative optimization. Experimental results on three heterogeneous benchmarks show that HGC-RC achieves competitive accuracy under high compression together with substantially reduced condensation cost. Future work includes evaluating broader backbone generalization and improving robustness under noisy structures and more challenging real-world heterogeneous settings.

\FloatBarrier
\bibliographystyle{IEEEtran}  % 指定IEEE风格
\bibliography{reference}      % 指定你的bib文件名为reference.bib

@inproceedings{GCN,
  author    = {Thomas N. Kipf and
               Max Welling},
  title     = {Semi-Supervised Classification with Graph Convolutional Networks},
  booktitle = {5th International Conference on Learning Representations, {ICLR} 2017},
  year      = {2017},
}

@inproceedings{GAT,
  author    = {Petar Velickovic and
               Guillem Cucurull and
               Arantxa Casanova and
               Adriana Romero and
               Pietro Li{\`{o}} and
               Yoshua Bengio},
  title     = {Graph Attention Networks},
  booktitle = {6th International Conference on Learning Representations, {ICLR} 2018},
  year      = {2018},
}

@inproceedings{HAN,
  author    = {Xiao Wang and
               Houye Ji and
               Chuan Shi and
               Bai Wang and
               Yanfang Ye and
               Peng Cui and
               Philip S. Yu},
  title     = {Heterogeneous Graph Attention Network},
  booktitle = {The World Wide Web Conference, {WWW} 2019},
  pages     = {2022--2032},
  year      = {2019},
}

@inproceedings{HGB,
  author    = {Qingsong Lv and
               Ming Ding and
               Qiang Liu and
               Yuxiang Chen and
               Wenzheng Feng and
               Siming He and
               Chang Zhou and
               Jianguo Jiang and
               Yuxiao Dong and
               Jie Tang},
  title     = {Are we really making much progress?: Revisiting, benchmarking and
               refining heterogeneous graph neural networks},
  booktitle = {{KDD} '21: The 27th {ACM} {SIGKDD} Conference on Knowledge Discovery and Data Mining},
  pages     = {1150--1160},
  year      = {2021},
}

@article{Nars,
  title={Scalable graph neural networks for heterogeneous graphs},
  author={Yu, Lingfan and Shen, Jiajun and Li, Jinyang and Lerer, Adam},
  journal={arXiv preprint arXiv:2011.09679},
  year={2020}
}

@inproceedings{SeHGNN,
  author       = {Xiaocheng Yang and
                  Mingyu Yan and
                  Shirui Pan and
                  Xiaochun Ye and
                  Dongrui Fan},
  title        = {Simple and Efficient Heterogeneous Graph Neural Network},
  booktitle    = {Thirty-Seventh {AAAI} Conference on Artificial Intelligence, 2023.},
  pages        = {10816--10824},
  year         = {2023},
}

@inproceedings{GraphSAGE,
  author    = {William L. Hamilton and
               Zhitao Ying and
               Jure Leskovec},
  title     = {Inductive Representation Learning on Large Graphs},
  booktitle = {Advances in Neural Information Processing Systems 30: Annual Conference on Neural Information Processing Systems 2017},
  pages     = {1024--1034},
  year      = {2017},
}

@inproceedings{Fast-GCN,
  author    = {Jie Chen and
               Tengfei Ma and
               Cao Xiao},
  title     = {FastGCN: Fast Learning with Graph Convolutional Networks via Importance
               Sampling},
  booktitle = {6th International Conference on Learning Representations, {ICLR} 2018},
  year      = {2018},
}

@inproceedings{GraphSAINT,
  author    = {Hanqing Zeng and
               Hongkuan Zhou and
               Ajitesh Srivastava and
               Rajgopal Kannan and
               Viktor K. Prasanna},
  title     = {GraphSAINT: Graph Sampling Based Inductive Learning Method},
  booktitle = {8th International Conference on Learning Representations, {ICLR}},
  year      = {2020},
}

@inproceedings{HetSANN,
  author       = {Huiting Hong and
                  Hantao Guo and
                  Yucheng Lin and
                  Xiaoqing Yang and
                  Zang Li and
                  Jieping Ye},
  title        = {An Attention-Based Graph Neural Network for Heterogeneous Structural
                  Learning},
  booktitle    = {The Thirty-Fourth {AAAI} Conference on Artificial Intelligence, {AAAI}
                  2020},
  pages        = {4132--4139},
  year         = {2020},
}

@inproceedings{HetGNN,  
author={Zhang, Chuxu and Song, Dongjin and Huang, Chao and Swami, Ananthram and Chawla, Nitesh V.}, 
 title={Heterogeneous Graph Neural Network}, 
 booktitle={Proceedings of the 25th ACM SIGKDD International Conference on Knowledge Discovery \& Data Mining}, 
 year={2019}, 
 }

@inproceedings{MAGNN,
  author       = {Xinyu Fu and
                  Jiani Zhang and
                  Ziqiao Meng and
                  Irwin King},
  title        = {{MAGNN:} Metapath Aggregated Graph Neural Network for Heterogeneous
                  Graph Embedding},
  booktitle    = {{WWW} '20: The Web Conference 2020},
  pages        = {2331--2341},
  year         = {2020},
}

@inproceedings{gcond,
  author       = {Wei Jin and
                  Lingxiao Zhao and
                  Shichang Zhang and
                  Yozen Liu and
                  Jiliang Tang and
                  Neil Shah},
  title        = {Graph Condensation for Graph Neural Networks},
  booktitle    = {The Tenth International Conference on Learning Representations, {ICLR}
                  2022, Virtual Event, April 25-29, 2022},
  year         = {2022},
}

@inproceedings{one-step,
  author       = {Wei Jin and
                  Xianfeng Tang and
                  Haoming Jiang and
                  Zheng Li and
                  Danqing Zhang and
                  Jiliang Tang and
                  Bing Yin},
  title        = {Condensing Graphs via One-Step Gradient Matching},
  booktitle    = {{KDD} '22: The 28th {ACM} {SIGKDD} Conference on Knowledge Discovery
                  and Data Mining},
  pages        = {720--730},
  year         = {2022},
}

@inproceedings{SFGC,
  author       = {Xin Zheng and
                  Miao Zhang and
                  Chunyang Chen and
                  Quoc Viet Hung Nguyen and
                  Xingquan Zhu and
                  Shirui Pan},
  title        = {Structure-free Graph Condensation: From Large-scale Graphs to Condensed Graph-free Data},
  booktitle    = {Advances in Neural Information Processing Systems 36: Annual Conference on Neural Information Processing Systems},
  year         = {2023},
}

@article{Receptive-Field-Distribution-Matching,
  author       = {Mengyang Liu and
                  Shanchuan Li and
                  Xinshi Chen and
                  Le Song},
  title        = {Graph Condensation via Receptive Field Distribution Matching},
  journal      = {CoRR},
  volume       = {abs/2206.13697},
  year         = {2022},
}

@article{Eigenbasis-Matching,
  author       = {Yang Liu and
                  Deyu Bo and
                  Chuan Shi},
  title        = {Graph Condensation via Eigenbasis Matching},
  journal      = {CoRR},
  volume       = {abs/2310.09202},
  year         = {2023},
}

@inproceedings{Dataset-Condensation1,
  author       = {George Cazenavette and
                  Tongzhou Wang and
                  Antonio Torralba and
                  Alexei A. Efros and
                  Jun{-}Yan Zhu},
  title        = {Dataset Distillation by Matching Training Trajectories},
  booktitle    = {{IEEE/CVF} Conference on Computer Vision and Pattern Recognition,
                  {CVPR} 2022, New Orleans, LA, USA, June 18-24, 2022},
  pages        = {10708--10717},
  year         = {2022},
}

@article{Dataset-Condensation2,
  author       = {Tongzhou Wang and
                  Jun{-}Yan Zhu and
                  Antonio Torralba and
                  Alexei A. Efros},
  title        = {Dataset Distillation},
  journal      = {CoRR},
  volume       = {abs/1811.10959},
  year         = {2018},
}

@inproceedings{Dataset-Condensation4,
  author       = {Bo Zhao and
                  Hakan Bilen},
  title        = {Dataset Condensation with Distribution Matching},
  booktitle    = {{IEEE/CVF} Winter Conference on Applications of Computer Vision, {WACV}
                  2023, Waikoloa, HI, USA, January 2-7, 2023},
  pages        = {6503--6512},
  year         = {2023},
}

@article{Sparsification1,
  author       = {Daniel A. Spielman and
                  Shang{-}Hua Teng},
  title        = {Spectral Sparsification of Graphs},
  journal      = {{SIAM} J. Comput.},
  volume       = {40},
  number       = {4},
  pages        = {981--1025},
  year         = {2011},
}

@inproceedings{HSparsification,
  author       = {Chandan Chunduru and
                  Chun Jiang Zhu and
                  Blake Gains and
                  Jinbo Bi},
  title        = {Heterogeneous Graph Sparsification for Efficient Representation Learning},
  booktitle    = {{IEEE} International Conference on Bioinformatics and Biomedicine,
                  {BIBM} 2022, Las Vegas, NV, USA, December 6-8, 2022},
  pages        = {1891--1896},
  year         = {2022},
}

@inproceedings{coreset1,
  author       = {Max Welling},
  title        = {Herding dynamical weights to learn},
  booktitle    = {Proceedings of the 26th Annual International Conference on Machine
                  Learning, {ICML} 2009, Montreal, Quebec, Canada, June 14-18, 2009},
  series       = {{ACM} International Conference Proceeding Series},
  volume       = {382},
  pages        = {1121--1128},
  year         = {2009},
}

@article{coreset2,
  author       = {Gert W. Wolf},
  title        = {Facility location: concepts, models, algorithms and case studies.
                  Series: Contributions to Management Science},
  journal      = {Int. J. Geogr. Inf. Sci.},
  volume       = {25},
  number       = {2},
  pages        = {331--333},
  year         = {2011},
}

@article{HGCond,
  author       = {Jian Gao and
                  Jianshe Wu and
                  Jingyi Ding},
  title        = {Heterogeneous Graph Condensation},
  journal      = {{IEEE} Trans. Knowl. Data Eng.},
  volume       = {36},
  number       = {7},
  pages        = {3126--3138},
  year         = {2024},
}

@inproceedings{GCEM, 
  author       = {Liu Yang and Deyu Bo and Chuan Shi},
  title        = {Graph Distillation with Eigenbasis Matching},
  booktitle    = {Proceedings of the International Conference on Machine Learning.},
  year         = {2024},
}

@inproceedings{graph-skeleton,
  author       = {Linfeng Cao and
                  Haoran Deng and
                  Yang Yang and
                  Chunping Wang and
                  Lei Chen},
  title        = {Graph-Skeleton: {\textasciitilde}1{\%} Nodes are Sufficient to Represent
                  Billion-Scale Graph},
  booktitle    = {Proceedings of the {ACM} on Web Conference 2024, {WWW}},
  pages        = {570--581},
  year         = {2024},
}

@article{survey-graph-reduction,
  author       = {Mohammad Hashemi and
                  Shengbo Gong and
                  Juntong Ni and
                  Wenqi Fan and
                  B. Aditya Prakash and
                  Wei Jin},
  title        = {A Comprehensive Survey on Graph Reduction: Sparsification, Coarsening,
                  and Condensation},
  journal      = {CoRR},
  volume       = {abs/2402.03358},
  year         = {2024},
}

@inproceedings{graph-coarsening,
  author       = {Zengfeng Huang and
                  Shengzhong Zhang and
                  Chong Xi and
                  Tang Liu and
                  Min Zhou},
  title        = {Scaling Up Graph Neural Networks Via Graph Coarsening},
  booktitle    = {{KDD} '21: The 27th {ACM} {SIGKDD} Conference on Knowledge Discovery
                  and Data Mining, Virtual Event, Singapore, August 14-18, 2021},
  pages        = {675--684},
  year         = {2021},
}

@inproceedings{gao1,
  title={Graph condensation for open-world graph learning},
  author={Gao, Xinyi and Chen, Tong and Zhang, Wentao and Li, Yayong and Sun, Xiangguo and Yin, Hongzhi},
  booktitle={Proceedings of the 30th ACM SIGKDD Conference on Knowledge Discovery and Data Mining},
  pages={851--862},
  year={2024}
}

@article{gao2,
  title={Rethinking and Accelerating Graph Condensation: A Training-Free Approach with Class Partition},
  author={Gao, Xinyi and Chen, Tong and Zhang, Wentao and Yu, Junliang and Ye, Guanhua and Nguyen, Quoc Viet Hung and Yin, Hongzhi},
  journal={arXiv preprint arXiv:2405.13707},
  year={2024}
}

@article{gao3,
  title={RobGC: Towards Robust Graph Condensation},
  author={Gao, Xinyi and Yin, Hongzhi and Chen, Tong and Ye, Guanhua and Zhang, Wentao and Cui, Bin},
  journal={arXiv preprint arXiv:2406.13200},
  year={2024}
}

@article{dse1,
  title={Poskhg: A position-aware knowledge hypergraph model for link prediction},
  author={Chen, Zirui and Wang, Xin and Wang, Chenxu and Li, Zhao},
  journal={Data Science and Engineering},
  volume={8},
  number={2},
  pages={135--145},
  year={2023},
}

@article{dse2,
  title={Few-shot relation prediction of knowledge graph via convolutional neural network with self-attention},
  author={Zhong, Shanna and Wang, Jiahui and Yue, Kun and Duan, Liang and Sun, Zhengbao and Fang, Yan},
  journal={Data Science and Engineering},
  volume={8},
  number={4},
  pages={385--395},
  year={2023},
}

@article{scis1,
  title={A novel graph oversampling framework for node classification in class-imbalanced graphs},
  author={Xia, Riting and Zhang, Chunxu and Zhang, Yan and Liu, Xueyan and Yang, Bo},
  journal={Science China Information Sciences},
  volume={67},
  number={6},
  pages={1--16},
  year={2024},
}

@article{scis2,
  title={Multiple types of disease-associated RNAs identification for disease prognosis and therapy using heterogeneous graph learning},
  author={Zhang, Wenxiang and Wei, Hang and Zhang, Wenjing and Wu, Hao and Liu, Bin},
  journal={Science China Information Sciences},
  volume={67},
  number={8},
  pages={189103},
  year={2024},
}

@article{jcst1,
  title={Label-Aware Chinese Event Detection with Heterogeneous Graph Attention Network},
  author={Cui, Shi-Yao and Yu, Bo-Wen and Cong, Xin and Liu, Ting-Wen and Tan, Qing-Feng and Shi, Jin-Qiao},
  journal={Journal of Computer Science and Technology},
  volume={39},
  number={1},
  pages={227--242},
  year={2024},
}

@article{jcst2,
  title={Meta-learning based Few-shot Link Prediction for Emerging Knowledge Graph},
  author={Zhang, Yu-Feng and Chen, Wei and Zhao, Peng-Peng and Xu, Jia-Jie and Fang, Jun-Hua and Zhao, Lei},
  journal={Journal of Computer Science and Technology},
  volume={39},
  number={5},
  pages={1058--1077},
  year={2024},
}

@inproceedings{
    gcpa,
    title={Adapting Precomputed Features for Efficient Graph Condensation},
    author={Yuan Li and Jun Hu and Zemin Liu and Bryan Hooi and Jia Chen and Bingsheng He},
    booktitle={Forty-second International Conference on Machine Learning},
    year={2025},
    url={https://openreview.net/forum?id=ThK6o74QLc}
}

@inproceedings{
Bonsai,
title={Bonsai: Gradient-free Graph Condensation for Node Classification},
author={Mridul Gupta and Samyak Jain and Vansh Ramani and Hariprasad Kodamana and Sayan Ranu},
booktitle={The Thirteenth International Conference on Learning Representations},
year={2025},
url={https://openreview.net/forum?id=5x88lQ2MsH}
}

@article{GCGP,
  title={Efficient Graph Condensation via Gaussian Process},
  author={Wang, Lin and Li, Qing},
  journal={arXiv preprint arXiv:2501.02565},
  year={2025}
}

@article{yuan2026novel,
  title={A Novel Approach to Temporal QoS Estimation via Extended Kalman Filter-Incorporated Latent Feature Analysis},
  author={Yuan, Ye and Wang, Song and Zhou, Hongxun and Wang, Ling and Luo, Xin},
  journal={IEEE Transactions on Services Computing},
  year={2026},
  doi={10.1109/TSC.2026.3697552}
}

@article{li2026adaptive,
  title={Adaptive PID-Incorporated Nonnegative Latent Factor Analysis},
  author={Li, Jinli and Yuan, Ye and He, Tiantian and Luo, Xin},
  journal={IEEE Transactions on Systems, Man, and Cybernetics: Systems},
  year={2026},
  doi={10.1109/TSMC.2026.3678292}
}

@article{wang2026graph,
  title={Graph Tensor Convolutional Network},
  author={Wang, Ling and Yuan, Ye and Luo, Xin},
  journal={IEEE Transactions on Systems, Man, and Cybernetics: Systems},
  year={2026},
  doi={10.1109/TSMC.2026.3655418}
}

@article{wang2026advanced,
  title={Advanced High-Order Graph Convolutional Networks with Assorted Time-Frequency Transforms},
  author={Wang, Ling and Yuan, Ye and Luo, Xin},
  journal={IEEE/CAA Journal of Automatica Sinica},
  volume={13},
  number={2},
  pages={394--408},
  year={2026}
}

@inproceedings{han2025sgd,
  title={SGD-DyG: Self-Reliant Global Dependency Apprehending on Dynamic Graphs},
  author={Han, Minglian and Wang, Ling and Yuan, Ye and Luo, Xin},
  booktitle={ACM SIGKDD Conference on Knowledge Discovery and Data Mining},
  pages={802--813},
  year={2025}
}

@article{li2025learning,
  title={Learning Error Refinement in Stochastic Gradient Descent-based Latent Factor Analysis via Diversified PID Controllers},
  author={Li, Jinli and Yuan, Ye and Luo, Xin},
  journal={IEEE Transactions on Emerging Topics in Computational Intelligence},
  volume={9},
  number={5},
  pages={3582--3597},
  year={2025}
}

@article{yuan2025node,
  title={A Node-Collaboration-Informed Graph Convolutional Network for Highly Accurate Representation to Undirected Weighted Graph},
  author={Yuan, Ye and Wang, Ying and Luo, Xin},
  journal={IEEE Transactions on Neural Networks and Learning Systems},
  volume={36},
  number={6},
  pages={11507--11519},
  year={2025}
}

@article{yuan2025proportional,
  title={A Proportional Integral Controller-Enhanced Non-negative Latent Factor Analysis Model},
  author={Yuan, Ye and Lu, Siyang and Luo, Xin},
  journal={IEEE/CAA Journal of Automatica Sinica},
  volume={12},
  number={6},
  pages={1246--1259},
  year={2025}
}

@article{wang2025gt,
  title={GT-A2T: Graph Tensor Alliance Attention Network},
  author={Wang, Ling and Liu, Kechen and Yuan, Ye},
  journal={IEEE/CAA Journal of Automatica Sinica},
  volume={12},
  number={10},
  pages={2165--2167},
  year={2025}
}

@article{yuan2024fuzzy,
  title={A Fuzzy PID-Incorporated Stochastic Gradient Descent Algorithm for Fast and Accurate Latent Factor Analysis},
  author={Yuan, Ye and Li, Jinli and Luo, Xin},
  journal={IEEE Transactions on Fuzzy Systems},
  volume={32},
  number={7},
  pages={4049--4061},
  year={2024}
}

@article{yuan2024adaptive,
  title={Adaptive Divergence-based Non-negative Latent Factor Analysis of High-Dimensional and Incomplete Matrices from Industrial Applications},
  author={Yuan, Ye and Luo, Xin and Zhou, MengChu},
  journal={IEEE Transactions on Emerging Topics in Computational Intelligence},
  volume={8},
  number={2},
  pages={1209--1222},
  year={2024}
}

@article{yuan2023kalman,
  title={A Kalman-Filter-Incorporated Latent Factor Analysis Model for Temporally Dynamic Sparse Data},
  author={Yuan, Ye and Luo, Xin and Shang, Mingsheng and Wang, Zidong},
  journal={IEEE Transactions on Cybernetics},
  volume={53},
  number={9},
  pages={5788--5801},
  year={2023}
}

@article{yuan2023adaptive,
  title={An Adaptive Divergence-based Non-negative Latent Factor Model},
  author={Yuan, Ye and Wang, Renfang and Yuan, Guangxiao and Luo, Xin},
  journal={IEEE Transactions on Systems, Man, and Cybernetics: Systems},
  volume={53},
  number={10},
  pages={6475--6487},
  year={2023}
}

@article{yuan2022multilayered,
  title={A Multilayered-and-Randomized Latent Factor Model for High-Dimensional and Sparse Matrices},
  author={Yuan, Ye and He, Qiang and Luo, Xin and Shang, Mingsheng},
  journal={IEEE Transactions on Big Data},
  volume={8},
  number={3},
  pages={784--794},
  year={2022}
}

@inproceedings{yuan2020generalized,
  title={A Generalized and Fast-converging Non-negative Latent Factor Model for Predicting User Preferences in Recommender Systems},
  author={Yuan, Ye and Luo, Xin and Shang, Mingsheng and Wu, Di},
  booktitle={The Web Conference},
  pages={498--507},
  year={2020}
}

@inproceedings{yuan2020temporal,
  title={Temporal Web Service QoS Prediction via Kalman Filter-Incorporated Dynamic Latent Factor Analysis},
  author={Yuan, Ye and Shang, Mingsheng and Luo, Xin},
  booktitle={European Conference on Artificial Intelligence},
  pages={561--568},
  year={2020}
}

@article{wu2026nongradient,
  title={Non-Gradient Hash Factor Learning for High-Dimensional and Incomplete Data Representation Learning},
  author={Wu, Di and Li, Shihui and He, Yi and Luo, Xin and Gao, Xinbo},
  journal={IEEE Transactions on Pattern Analysis and Machine Intelligence},
  year={2026},
  doi={10.1109/TPAMI.2026.3653780}
}

@article{qin2026robust,
  title={A Robust Approach to Electricity Theft Detection via Tensor Representation-Driven Contrastive Distillation},
  author={Qin, Wen and Ding, Yuting and Luo, Xin},
  journal={IEEE Transactions on Industrial Informatics},
  year={2026},
  doi={10.1109/TII.2026.3659333}
}

@article{wu2025learning,
  title={Learning Accurate Representation to Nonstandard Tensors via a Mode-Aware Tucker Network},
  author={Wu, Hao and Wang, Qu and Luo, Xin and Wang, Zidong},
  journal={IEEE Transactions on Knowledge and Data Engineering},
  volume={37},
  number={12},
  pages={7272--7285},
  year={2025}
}

@article{lyu2026dynamic,
  title={Dynamic Stochastic Reorientation Particle Swarm Optimization for Adaptive Latent Factor Analysis in High-Dimensional Sparse Matrices},
  author={Lyu, Chao and Ma, Ziwen and Luo, Xin and Shi, Yuhui},
  journal={IEEE Transactions on Knowledge and Data Engineering},
  volume={38},
  number={1},
  pages={222--234},
  year={2026}
}

@article{he2026tensor,
  title={Tensor Low-Rank Orthogonal Compression for Convolutional Neural Networks},
  author={He, Yaping and Luo, Xin},
  journal={IEEE/CAA Journal of Automatica Sinica},
  volume={13},
  number={1},
  pages={227--229},
  year={2026}
}

@article{bi2025discovering,
  title={Discovering Spatio-Temporal-Individual Coupled Features from Nonstandard Tensors-A Novel Dynamic Graph Mixer Approach},
  author={Bi, Fanghui and He, Tiantian and Ong, Yew-Soon and Luo, Xin},
  journal={IEEE Transactions on Neural Networks and Learning Systems},
  volume={36},
  number={11},
  pages={19834--19848},
  year={2025}
}

@article{bi2025spatiotemporal,
  title={Spatiotemporal Graph Neural Network-Incorporated Latent Factorization of Tensors for Dynamic QoS Estimation},
  author={Bi, Fanghui and He, Tiantian and Luo, Xin},
  journal={IEEE/CAA Journal of Automatica Sinica},
  year={2025},
  doi={10.1109/JAS.2025.125750}
}

@article{he2025modularized,
  title={Modularized Graph Convolutional Network},
  author={He, Tiantian and Duan, Zhixuan and Luo, Xin},
  journal={IEEE/CAA Journal of Automatica Sinica},
  year={2025},
  doi={10.1109/JAS.2025.125336}
}

@article{tang2025autoencoding,
  title={Auto-Encoding Neural Tucker Factorization},
  author={Tang, Peng and Luo, Xin and Woodcock, Jim},
  journal={IEEE Transactions on Knowledge and Data Engineering},
  volume={37},
  number={10},
  pages={5795--5807},
  year={2025}
}

@article{lyu2025genetic,
  title={Genetic Algorithm-based Two-Step Optimization for Precise Latent Factor Analysis},
  author={Lyu, Chao and Cheng, Jingna and Luo, Xin and Shi, Yuhui},
  journal={IEEE Transactions on Neural Networks and Learning Systems},
  year={2025},
  doi={10.1109/TNNLS.2025.3631465}
}

@article{lin2026ncsac,
  title={NCSAC: Effective Neural Community Search via Attribute-augmented Conductance},
  author={Lin, Longlong and Li, Quanao and Qiao, Miao and Wang, Zeli and Zhao, Jin and Li, Rong-Hua and Luo, Xin and Jia, Tao},
  journal={IEEE Transactions on Knowledge and Data Engineering},
  volume={38},
  number={2},
  pages={1221--1235},
  year={2026}
}

@article{wu2025multi,
  title={Multi Metric Autoencoder for Representing High-Dimensional and Incomplete Data},
  author={Wu, Di and Liang, Cheng and He, Yi and Qiao, Yan and Luo, Xin},
  journal={IEEE Transactions on Systems, Man, and Cybernetics: Systems},
  year={2025},
  doi={10.1109/TSMC.2025.3646863}
}

@article{liao2025proximal,
  title={A Proximal-ADMM-incorporated Nonnegative Latent-Factorization-of-Tensors Model for Representing Dynamic Cryptocurrency Transaction Network},
  author={Liao, Xin and Wu, Hao and He, Tiantian and Luo, Xin},
  journal={IEEE Transactions on Systems, Man, and Cybernetics: Systems},
  volume={55},
  number={11},
  pages={8387--8401},
  year={2025}
}

@article{hu2025comprehensive,
  title={A Comprehensive Review of Parallel Optimization Algorithms for High-Dimensional and Incomplete Matrix Factorization},
  author={Hu, Qicong and Wu, Hao and Luo, Xin},
  journal={IEEE/CAA Journal of Automatica Sinica},
  volume={12},
  number={12},
  pages={2399--2426},
  year={2025}
}

@article{xu2026sampling,
  title={A Sampling-Neighborhood-Regularized Latent Factorization of Tensor for Dynamic QoS Estimation},
  author={Xu, Xiuqin and Lin, Mingwei and Xu, Zeshui and Luo, Xin},
  journal={IEEE Transactions on Network and Service Management},
  volume={23},
  pages={1707--1722},
  year={2026}
}

@article{liao2025novel,
  title={A Novel Tensor Causal Convolution Network Model for Highly-Accurate Representation to Spatio-Temporal Data},
  author={Liao, Xin and Wu, Hao and Luo, Xin},
  journal={IEEE Transactions on Automation Science and Engineering},
  volume={22},
  pages={19525--19537},
  year={2025}
}

@article{wang2025convolution,
  title={A Convolution Bias-Incorporated Nonnegative Latent Factorization of Tensors Model for Accurate Representation Learning to Dynamic Directed Graphs},
  author={Wang, Qu and Wu, Hao and Luo, Xin},
  journal={IEEE Transactions on Systems, Man, and Cybernetics: Systems},
  volume={55},
  number={12},
  pages={8902--8914},
  year={2025}
}

@article{xu2025attention,
  title={Attention-Mechanism-Based Neural Latent-Factorization-of-Tensors Mode},
  author={Xu, Xiuqin and Lin, Mingwei and Xu, Zeshui and Luo, Xin},
  journal={ACM Transactions on Knowledge Discovery from Data},
  volume={19},
  number={4},
  pages={1--27},
  year={2025}
}

@article{li2025knowledge,
  title={Knowledge-driven Multiple Instance Learning with Hierarchical Cluster-incorporated Aware Filtering for Larynx Pathological Grading},
  author={Li, Chentao and Huang, Pan and Qin, Jing and Luo, Xin},
  journal={IEEE Journal of Biomedical and Health Informatics},
  year={2025},
  doi={10.1109/JBHI.2025.3609838}
}

@article{yang2025link,
  title={Link-based Attributed Graph Clustering via Approximate Generative Bayesian Learning},
  author={Yang, Yue and Hu, Lun and Li, Guodong and Li, Dongxu and Hu, Pengwei and Luo, Xin},
  journal={IEEE Transactions on Systems, Man, and Cybernetics: Systems},
  volume={55},
  number={8},
  pages={5730--5743},
  year={2025}
}

@article{yang2025fmvpci,
  title={Fmvpci: A Multi-View Fusion Neural Network for Identifying Protein Complex via Fuzzy Clustering},
  author={Yang, Yue and Hu, Lun and Li, Guodong and Li, Dongxu and Hu, Pengwei and Luo, Xin},
  journal={IEEE Transactions on Systems, Man, and Cybernetics: Systems},
  volume={55},
  number={9},
  pages={6189--6202},
  year={2025}
}

@article{xu2025adaptively,
  title={An Adaptively Bias-Extended Non-negative Latent Factorization of Tensors Model for Accurately Representing the Dynamic QoS Data},
  author={Xu, Xiuqin and Lin, Mingwei and Luo, Xin and Xu, Zeshui},
  journal={IEEE Transactions on Services Computing},
  volume={18},
  number={2},
  pages={603--617},
  year={2025}
}

@article{wu2025graph,
  title={Graph-Based Prediction of miRNA-Drug Associations with Multisource Information and Metapath Enhancement Matrices},
  author={Wu, Ming-Yang and Hu, Pengwei and You, Zhu-Hong and Zhang, Jun and Hu, Lun and Luo, Xin},
  journal={IEEE Journal of Biomedical and Health Informatics},
  year={2025},
  doi={10.1109/JBHI.2025.3558303}
}

@article{hou2025multiaspect,
  title={Multi-Aspect Self-Attending Neural Tucker Factorization for Spatiotemporal Representation Learning},
  author={Hou, Yikai and Tang, Peng and Luo, Xin},
  journal={IEEE/CAA Journal of Automatica Sinica},
  year={2025},
  doi={10.1109/JAS.2025.125723}
}

@article{li2026neural,
  title={Neural Non-Negative Latent Factorization of Tensors Model with Acceleration and Unconstraint},
  author={Li, Wenqiang and Lin, Mingwei and Xu, Xiuqin and Lin, Ling and Xu, Zeshui and Luo, Xin},
  journal={IEEE Transactions on Systems, Man, and Cybernetics: Systems},
  volume={56},
  number={1},
  pages={164--178},
  year={2026}
}

@article{deng2026fuzzy,
  title={Fuzzy Mixture-of-Experts Aggregation for Organoid Identification with Multi-Scale State Space Features},
  author={Deng, Xun and Hu, Pengwei and Herget, Thomas and Tan, Feng and Zhu, Xiaobo and Zhang, Jun and Huang, Yu-an and Hu, Lun and You, Zhuhong and Luo, Xin},
  journal={IEEE Transactions on Fuzzy Systems},
  volume={34},
  number={1},
  pages={324--335},
  year={2026}
}

@article{gou2026multiscale,
  title={Multi-Scale Collaborative Distillation Graph Neural Networks for Session-Based Recommendation},
  author={Gou, Jianping and Cheng, Youhui and Ma, Benteng and Du, Lan and Luo, Xin and Yi, Zhang},
  journal={IEEE Transactions on Services Computing},
  volume={19},
  number={1},
  pages={504--517},
  year={2026}
}

@article{liu2025scalable,
  title={A Scalable Multi-Channel Sentiment Analysis Model with Enhanced Semantic Understanding and Redundancy Reduction},
  author={Liu, Jun and Li, Xiang and Lin, Mingwei and Luo, Xin},
  journal={IEEE Transactions on Computational Social Systems},
  year={2025},
  doi={10.1109/TCSS.2025.3619188}
}

@article{luo2025calibrator,
  title={A Calibrator Fuzzy Ensemble for Highly-Accurate Robot Arm Calibration},
  author={Luo, Xin and Li, Zhibin and Yue, Wenbin and Li, Shuai},
  journal={IEEE Transactions on Neural Networks and Learning Systems},
  volume={36},
  number={2},
  pages={2169--2181},
  year={2025}
}

@article{luo2025analysis,
  title={Analysis of Students’ Positive Emotion and Smile Intensity Using Sequence-Relative Key-Frame Labeling and Deep-Asymmetric Convolutional Neural Network},
  author={Luo, Zhenzhen and Jin, Xiaolu and Luo, Yong and Zhou, Qiangqiang and Luo, Xin},
  journal={IEEE/CAA Journal of Automatica Sinica},
  volume={12},
  number={4},
  pages={806--820},
  year={2025}
}

@article{bi2025graph,
  title={Graph Linear Convolution Pooling for Learning in Incomplete High-Dimensional Data},
  author={Bi, Fanghui and He, Tiantian and Ong, Yew-Soon and Luo, Xin},
  journal={IEEE Transactions on Knowledge and Data Engineering},
  volume={37},
  number={4},
  pages={1838--1852},
  year={2025}
}

@article{he2025structure,
  title={Structure-Preserved Self-Attention for Fusion Image Information in Multiple Color Spaces},
  author={He, Zhu and Lin, Mingwei and Luo, Xin and Xu, Zeshui},
  journal={IEEE Transactions on Neural Networks and Learning Systems},
  volume={36},
  number={7},
  pages={13021--13035},
  year={2025}
}

@article{wu2025outlier,
  title={An Outlier-Resilient Autoencoder for Representing High-Dimensional and Incomplete Data},
  author={Wu, Di and Hu, Yuanpeng and Liu, Kechen and Li, Jing and Wang, Xianmin and Deng, Song and Zheng, Nenggan and Luo, Xin},
  journal={IEEE Transactions on Emerging Topics in Computational Intelligence},
  volume={9},
  number={2},
  pages={1379--1391},
  year={2025}
}

@article{chen2025latent,
  title={Latent Factorization of Tensors Incorporated Battery Cycle Life Prediction},
  author={Chen, Minzhi and Tao, Li and Lou, Jungang and Luo, Xin},
  journal={IEEE/CAA Journal of Automatica Sinica},
  volume={12},
  number={3},
  pages={633--635},
  year={2025}
}

@article{qin2024adaptively,
  title={Adaptively-accelerated Parallel Stochastic Gradient Descent for High-Dimensional and Incomplete Data Representation Learning},
  author={Qin, Wen and Luo, Xin and Zhou, MengChu},
  journal={IEEE Transactions on Big Data},
  volume={10},
  number={1},
  pages={92--107},
  year={2024}
}

@article{li2024generalized,
  title={A Generalized Nesterov-Accelerated Second-Order Latent Factor Model for High-Dimensional and Incomplete Data},
  author={Li, Weiling and Wang, Renfang and Luo, Xin},
  journal={IEEE Transactions on Neural Networks and Learning Systems},
  volume={36},
  number={1},
  pages={1518--1532},
  year={2024}
}

@article{bi2024fast,
  title={A Fast Nonnegative Autoencoder-based Approach to Latent Feature Analysis on High-Dimensional and Incomplete Data},
  author={Bi, Fanghui and He, Tiantian and Luo, Xin},
  journal={IEEE Transactions on Services Computing},
  volume={17},
  number={3},
  pages={733--746},
  year={2024}
}

@article{qin2024asynchronous,
  title={Asynchronous Parallel Fuzzy Stochastic Gradient Descent for High-Dimensional Incomplete Data},
  author={Qin, Wen and Luo, Xin},
  journal={IEEE Transactions on Fuzzy Systems},
  volume={32},
  number={2},
  pages={445--459},
  year={2024}
}

@article{liu2024symmetry,
  title={Symmetry and Graph Bi-regularized Non-Negative Matrix Factorization for Precise Community Detection},
  author={Liu, Zhigang and Luo, Xin and Zhou, MengChu},
  journal={IEEE Transactions on Automation Science and Engineering},
  volume={21},
  number={2},
  pages={1406--1420},
  year={2024}
}

@article{wu2024prediction,
  title={A Prediction-sampling-based Multilayer-structured Latent Factor Model for Accurate Representation to High-dimensional and Sparse Data},
  author={Wu, Di and Luo, Xin and He, Yi and Zhou, MengChu},
  journal={IEEE Transactions on Neural Networks and Learning Systems},
  volume={35},
  number={3},
  pages={3845--3858},
  year={2024}
}

@article{chen2024generalized,
  title={A Generalized Nesterov's Accelerated Gradient-Incorporated Non-negative Latent-factorization-of-tensors Model for Efficient Representation to Dynamic QoS Data},
  author={Chen, Minzhi and Qiao, Yan and Wang, Renfang and Luo, Xin},
  journal={IEEE Transactions on Emerging Topics in Computational Intelligence},
  volume={8},
  number={3},
  pages={2386--2400},
  year={2024}
}
\end{document}